# MapperGPT: Large Language Models for Linking and Mapping Entities

Nicolas Matentzoglu[1], J. Harry Caufield[2], Harshad B. Hegde[2], Justin T. Reese[2], Sierra Moxon[2], Hyeongsik Kim[3], Nomi L. Harris[2], Melissa A Haendel[4], Christopher J. Mungall[2*]

[1]Semanticly Ltd., Athens, Greece
[2]Lawrence Berkeley National Laboratory, Berkeley, CA 94720, USA
[3]Robert Bosch LLC
[4]Anschutz Medical Campus, Aurora, CO 80045, USA, University of Colorado

* Corresponding author, cjmungall@lbl.gov

## Abstract

Aligning terminological resources, including ontologies, controlled vocabularies, taxonomies, and value sets is a critical part of data integration in many domains such as healthcare, chemistry, and biomedical research. Entity mapping is the process of determining correspondences between entities across these resources, such as gene identifiers, disease concepts, or chemical entity identifiers. Many tools have been developed to compute such mappings based on common structural features and lexical information such as labels and synonyms. Lexical approaches in particular often provide very high recall, but low precision, due to lexical ambiguity. As a consequence of this, mapping efforts often resort to a labor intensive manual mapping refinement through a human curator.

Large Language Models (LLMs), such as the ones employed by ChatGPT, have generalizable abilities to perform a wide range of tasks, including question-answering and information extraction. Here we present *MapperGPT*, an approach that uses LLMs to review and refine mapping relationships as a post-processing step, in concert with existing high-recall methods that are based on lexical and structural heuristics.

We evaluated MapperGPT on a series of alignment tasks from different domains, including anatomy, developmental biology, and renal diseases. We devised a collection of tasks that are designed to be particularly challenging for lexical methods. We show that when used in combination with high-recall methods, MapperGPT can provide a substantial improvement in accuracy, beating state-of-the-art (SOTA) methods such as LogMap.

## Introduction

Tackling difficult challenges in the area of biological and biomedical research, such as rare disease diagnostics and variant prioritization, requires the integration of a large number of disparate data sources. Due to the decentralized nature of data standardization, where different data providers inevitably employ different controlled vocabularies and ontologies to standardize their data, and aggravated by the need to link data across different species with strongly diverging terminologies, it becomes crucial to integrate such "semantic spaces" (i.e. data spaces that are described using divergent sets of ontologies) by linking entities. For example, integrating genetic associations for a disease provided by a disease data resource such as the Online Mendelian Inheritance in Man (OMIM, https://omim.org/) with the phenotypic associations to the same disease provided by Orphanet (https://www.orpha.net/) requires mapping different disease identifiers that refer to the exact same real-world disease concept. Manually mapping thousands of disease concepts between two semantic spaces is potentially realizable, but in the real world, dozens of resources providing information about the same data type (disease, genes, environment, organisms) need to be integrated, which makes a purely manual approach prone to substantial omissions and/or significant lag time.

Semantic entity matching is the process of associating a term *A* (we use "term" to mean the concept encoded by an "identifier" in the following) in one semantic space to a term *B* in another, where A and B refer to the same or related real-world concepts. A common way to automate semantic entity matching is to use lexical methods, in particular matching on primary or alternative labels (synonyms) that have been assigned to concepts, sometimes in combination with lexical normalization (e.g. lower-casing or lemmatization). These methods can often provide very high recall, but low precision, due to lexical ambiguity. Examples are provided in Table 1, including a false positive match.

| Resource A | Concept A | Resource B | Concept B | Predicted relation | True relation |
| --- | --- | --- | --- | --- | --- |
| UK Auto Ontology | Car | Industrial Ontology | Automobile | n/a | exactMatch |
| Train Ontology | Car | Industrial Ontology | Railway Carriage | n/a | closeMatch |
| Gazetteer (GAZ) | Colon | Uberon Anatomy Ontology | colon | exactMatch | differentFrom |

**Table 1**: Examples of mappings generated by a lexical matching tool. The match between Colon (the city in Panama) in GAZ and the anatomical part "colon" in the Uberon Anatomy ontology due to the equality of the class names is a false positive match.

An example of this approach is the LOOM algorithm used in the BioPortal ontology resource[1], which provides very high recall mappings across over a thousand ontologies and other controlled vocabularies. A number of other approaches, many of which make use of other relationships or properties in the ontology, can give higher precision mappings. The Ontology Alignment Evaluation Initiative (OAEI) provides a yearly evaluation of different methods for ontology matching[2]. One of the top-performing methods of OAEI is LogMap[3], which makes use of logical axioms (such as SubClassOf or DisjointWith) in the ontologies to assist in mapping.

Deep learning approaches and in particular Language Models (LMs) have been applied to ontology matching tasks. Some methods make use of embedding distance, such as

OntoEmma[4], DeepAlignment[5], VeeAlign[6]. More recently the Truveta Mapper[7] treats matching as a translation task and involves pre-training on ontology structures.

The most recent development in LMs are instruction-tuned Large Language Models (LLMs), exemplified by ChatGPT, which involve billions of parameters and pre-training on instruction-prompting tasks. The resulting models have generalizable abilities to perform a wide range of tasks, including question-answering and information extraction. He et al. (2023)[8] concluded that the use of LLMs for the ontology matching problem is "promising", but some challenges (such as prompt tuning and overall framework design) still need to be addressed.

Given their performance on any tasks related to the understanding and generation of natural language, it seems obvious that LLMs could be used directly as a powerful, scalable alternative to current SOTA methods for entity matching. One possibility is using LLMs like the ones employed by ChatGPT to generate mappings *de novo*. There are various challenges with asking the LLM directly to suggest a suitable mapping. Firstly the training set cutoff is often many months or even years in the past (ChatGPT cutoff is January 2022 at the time of this writing (October 2023)) which means that the LLM very likely does not know about the latest state of the resources being mapped (if it knows them at all). Secondly LLMs all by themselves are prone to hallucination, in particular, when it comes to the generation of database or ontology identifiers. To compensate for the training cutoff and, at least to some extent, the hallucination, the resource and its identifiers could be provided "in context", i.e. provided as part of the prompt. Given the ontologies' (frequently) large size this is currently not possible due to severe limitations in window size in most current LLMs. GPT4, for example, currently has a limited window size of 32,000 tokens, which corresponds roughly to 128,000 characters, while a typical biomedical ontology has millions of characters (Uberon, for example, has more than 18 million). A third option could be to feed the powerset of all possible pairs of mappings to the LLM. This is, given the size of the powerset (a moderately sized biomedical ontology can have around 10,000 classes, which adds up to 100,000,000 pairwise comparisons) both prohibitively costly from a runtime perspective (current LLMs are slow, with GPT4 being able to generate around 5 tokens of generated output per second) and also from a financial standpoint.

We devised an alternative approach called *MapperGPT* that does not use GPT models to generate mappings de-novo, but instead works in concert with existing high-recall methods such as LOOM[1]. The method expects a set of candidate mappings with potentially numerous false positives as an input, and then uses a GPT model to review and refine those mappings as a post-processing step, essentially for the purpose of isolating and removing false positive mappings. We use an approach in which examples of different mapping categories and information about the two concepts in a mapping are provided to the model to determine an appropriate mapping relationship. We use the Simple Standard for Sharing Ontological Mapping (SSSOM)[9] for sharing and comparing entity mappings across systems. The overall approach thereby refines domain knowledge-derived mappings using a language model's semantic representations.

We evaluated MapperGPT on a series of alignment tasks from different domains, including anatomy, developmental biology, and renal diseases. We devised a collection of tasks that are designed to be particularly challenging for lexical methods, where naive lexical matching can lead to false positives due to terminological clashes. We show that when used in combination with high-recall methods such as LOOM or OAK Lexmatch (https://github.com/INCATools/ontology-access-kit), MapperGPT can provide a substantial improvement in accuracy, beating SOTA methods such as LogMap.

Our key contributions are as follows:
- An algorithm and tool, MapperGPT, that uses a GPT model to review and refine mapping relationships between terms
- A collection of new matching tasks expressed using the SSSOM standard, which we used to test and tune MapperGPT and can be reused to benchmark other mapping tools

## Methods

*Algorithm.* Our method MapperGPT takes as input two ontologies O1 and O2 and a set of candidate mappings M. These mappings are assumed to have been generated from an existing high-recall method such as LOOM. MapperGPT starts by generating a prompt for each candidate mapping in the input set:

```
M' = {}
for m in M:
  prompt = GeneratePrompt(m.a, m.b, O1, O2)
  response = CompletePrompt(prompt, model)
  m' = Parse(response)
  add m' to M'
return M'
```

*Prompt Generation.* The method GeneratePrompt generates a prompt according to the following template:

```
What is the relationship between the two specified concepts?

Give your answer in the form:

category: <one of: EXACT_MATCH, BROADER_THAN, NARROWER_THAN, RELATED_TO, DIFFERENT>
confidence: <one of: LOW, HIGH, MEDIUM>
similarities: <semicolon-separated list of similarities>
differences: <semicolon-separated list of differences>

Make use of all provided information, including the concept names, definitions, and relationships.

Examples:
```

```
{{ examples }}

Here are the two concepts:

{{ Describe(conceptA) }}
{{ Describe(conceptB) }}
```

Concepts and the properties of their mappings are provided as part of the prompt with their expected structure and level of detail. Examples are provided in-context in the following form:

```
[Concept A]
id: FOO:125
name: wing
def: part of a bird that is flapped to enable flight
is_a: Limb
relationship: part_of Bird
relationship: has_part Feather

[Concept B]
id: BAR:458
name: wing
relationship: part_of Aeroplane

category: DIFFERENT
confidence: HIGH
similarities: function
differences: A is an anatomical part; B is a part of a vehicle
```

For each candidate mapping between concepts A and B, we generate a description of each concept, incorporating key elements: name, synonyms, definition, relationships. The Describe function generates a textual description of an ontology or database concept, showing the following properties:
- name
- synonyms
- definition
- parents (superclasses)
- other relationships

*Prompt Completion.* The prompt is passed to a GPT model, which generates a response. In principle the method should work with any instruction-based model, either local or remotely accessed via an API. In practice we have only evaluated this against the OpenAI API and the two leading instruction-based models, gpt-3.5-turbo and gpt-4.

*Response Parsing.* The response is parsed to retrieve the key data model elements: category, confidence, similarities, and differences. The result object can be exported to SSSOM format.

*As an example*, two concepts from the Drosophila (fruit fly) and zebrafish anatomy ontologies[10,11] are candidate matches due to sharing a lexical element (the "PC" abbreviation). This is a false positive match in reality, as the concepts are entirely different. The two concept descriptions are generated from respective ontologies as follows:

```
[Concept A]
id: FBbt:00001906
name: embryonic/larval Malpighian tubule Type I cell
def: Type I cell of the embryonic/larval Malpighian tubules.
synonyms:  PC ;  embryonic/larval Malpighian tubule Type I cell ;
larval Malpighian tubule Type I cell ;  larval Malpighian tubule
principal cell ;
is_a:  embryonic/larval specialized Malpighian tubule cell ;  Malpighian
tubule Type I cell ;

[Concept B]
id: ZFA:0000320
name: caudal commissure
def: Diencephalic tract which is located in the vicinity of the dorsal
diencephalon and mesencephalon and connects the pretectal nuclei. From
Neuroanatomy of the Zebrafish Brain.
synonyms:  PC ;  caudal commissure ;  posterior commissure ;
is_a:  diencephalic white matter ;
relationship: part of synencephalon
relationship: start stage unknown
relationship: end stage adult
```

The response for this using gpt-3.5-turbo (August 2023) is:

```
category: DIFFERENT
confidence: HIGH
similarities: NONE
differences: A is a type of cell in the embryonic/larval Malpighian
tubules; B is a diencephalic tract in the zebrafish brain.
subject: FBbt:00001906
object: ZFA:0000320
This is then parsed to a YAML object:
predicate: DIFFERENT
confidence: HIGH
similarities:
  - NONE
```

```
differences:
  - A is a type of cell in the embryonic/larval Malpighian tubules
  - B is a diencephalic tract in the zebrafish brain.
```

The consumer may typically only make use of the predicate slot, but the list of similarities and differences may prove informative.

## Implementation

We use the Ontology Access Kit (OAK) library[12] to connect to a variety of ontologies in the Open Biological and Biomedical Ontology (OBO) Foundry[13] and BioPortal[14]. OAK provides general access to ontologies, but we also make use of its ability to extract subsets of ontologies, perform lexical matching using OAK Lexmatch, extract mappings from ontologies and ontology portals such as BioPortal, and add labels to mapping tables which typically only include the mapped identifiers, for better readability. We make use of ROBOT[15] for converting between different ontology formats. The overall mapping framework is implemented in OntoGPT[16] (https://github.com/monarch-initiative/ontogpt) in a method called categorize-mappings, where the input is a SSSOM mapping file (usually generated by a lexical matching tool) and the output is a SSSOM mapping file with predicate_id filled with the predicted value. We call OntoGPT from the command line like this:

```
ontogpt categorize-mappings --model gpt-4 -i foo.sssom.tsv -o bar.sssom.tsv
```

All test sets and generated mappings sets are available online (https://github.com/monarch-initiative/gpt-mapping-manuscript). The entire pipeline is implemented as a fully reproducible Makefile.

## Generation of test sets

To evaluate our approach, we created a collection of test sets from the biological and biomedical domains. We chose to devise new test sets as we wanted to base these on up-to-date, precise, validated mappings derived from ontologies such as Mondo[17], Cell Ontology (CL)[18], and the Uberon Anatomy Ontology (Uberon)[19].

To generate anatomy test sets (see Table 2), we generated pairwise mappings between species-specific anatomy ontologies, using the Uberon and CL mappings as the gold standard. If a pair of concepts are transitively linked via Uberon or CL, then they are considered a match. For example, UBERON:0000924 (ectoderm) is mapped to FBbt:00000111 (ectoderm (fruitfly)) and ZFA:0000016 (ectoderm (zebrafish)), so we assume that FBbt:00000111 is an exact match to ZFA:0000016. We used the same method for linking species-specific developmental stage terms.

We also generated a test set from a group of disease terms in Mondo (heritable renal diseases) and their curated, validated mappings to corresponding disease terms in NCIT.

| Test set | Size (skos:exactMatch) | Source |
|---|---|---|
| MONDO-NCIT renal subset | 25 | Mondo (curated mappings) |
| HSAPV-MMUSDV | 22 | Uberon (curated mappings) |
| FBbt-WBbt | 41 | Uberon (curated mappings) |
| FBbt-ZFA | 72 | Uberon (curated mappings) |

**Table 2**: Breakdown of the existing test sets.

## Tool evaluation

We evaluate MapperGPT with two models: gpt-3.5-turbo and gpt-4. MapperGPT is capable of providing refined predicates from SKOS beyond skos:exactMatch. For this task, however, we only determine exact matches, and discard all others, to ensure a more fair comparison with the other tools which are optimized for determining exact matches.

We also evaluated against the OAK Lexmatch tool, as a high-recall baseline. Although Lexmatch allows for customizable rules, we ran this without any prior knowledge of the domains, and considered any lexical match to be a predicted match.

We selected LogMap, which is one of the top-performing mappers in the OAEI as the SOTA method to compare MapperGPT against. We convert LogMap results to SSSOM format using the SSSOM toolkit (https://github.com/mapping-commons/sssom-py).

LogMap produces a score with each mapping, so we scanned all scores to determine the optimal score threshold in terms of accuracy (F1). Note that the fact that LogMap provides a score rather than making a committed choice gives LogMap an advantage over our method (which has to make a choice), because we have to use the test set to actually select the "optimal cutoff", i.e. the score at which we are confident that the mapping more frequently corresponds to the ground truth.

## Results

Here, we present the outcomes of our evaluations of the MapperGPT approach. Our method is intended to refine a set of overestimation mappings (e.g. those derived from a naive lexical matching tool) by leveraging a language model's semantic representations. Accordingly, we find that MapperGPT offers significant and remarkable improvements in refining pairwise mappings across our test sets. We reiterate that we designed these tasks to pose challenges to lexical methods, as overly simplistic lexical matching can yield false positives due to terminological conflicts. When used in conjunction with high-recall methods such as LOOM or OAK Lexmatch, MapperGPT yields a substantial improvement in mapping accuracy, surpassing LogMap results on the same sets.

## MapperGPT with GPT-4 improves on state of the art across all tasks

On all tasks combined, summarized in Table 3, MapperGPT with GPT-4 has an accuracy of 0.672 (as per F1 score). This constitutes a considerable improvement over the SOTA (LogMap with 0.527), ca. 24%, demonstrating the validity of the approach. A score distribution is provided in Figure 1. As expected, the baseline method (Lexmatch) achieves high recall but very low precision. Neither LogMap nor MapperGPT with GPT-3 yield higher recall though both demonstrate higher than baseline precision. MapperGPT with GPT-4, however, reaches comparatively high scores for both precision and recall.

| method | F1 | P | R |
|---|---|---|---|
| lexmatch | 0.340 | 0.210 | **0.881** |
| logmap | 0.527 | 0.458 | 0.619 |
| gpt3 | 0.490 | 0.500 | 0.481 |
| gpt4 | **0.672** | **0.601** | 0.762 |

**Table 3.** Results for all mapping tasks. On recall alone, the baseline Lexmatch approach performs best, but predictably suffers from low precision. As measured by F1 score, LogMap performs better than our MapperGPT approach only when GPT-3 is used. MapperGPT with GPT-4 achieves precision and recall higher than that of LogMap.

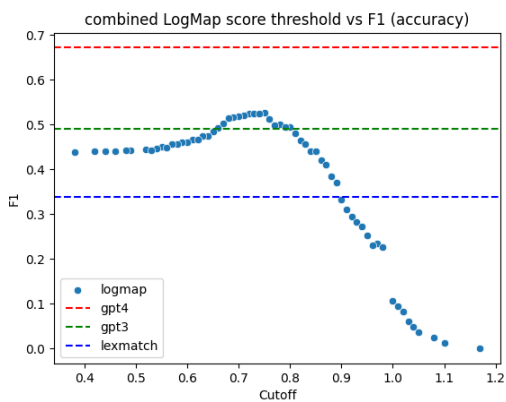

**Figure 1.** Combined LogMap score threshold vs. accuracy of other approaches across all mapping tasks. LogMap scores reach a plateau around a confidence threshold of 0.7 to 0.8 (for the evaluations in Table 1, a cutoff of 0.8 is used). While this cutoff places the overall F1 score for LogMap in the same range as that achieved by MapperGPT with GPT-3, MapperGPT with GPT-4 demonstrates accuracy exceeding that of LogMap at any threshold.

## Anatomy Task Results

In the first anatomy ontology matching task, methods were evaluated on their ability to reproduce all curated mappings between the Fly anatomy ontology (FBbt) and the Zebrafish anatomy ontology (ZFA). A full comparison of the results is provided in **Table 4**, with score distributions in **Figure 2**. Here, MapperGPT with GPT-4 demonstrates a higher F1 score than all other methods, though its precision is similar to that of our method using GPT-3.

| method | F1 | P | R |
|---|---|---|---|
| lexmatch | 0.349 | 0.219 | **0.847** |
| logmap | 0.486 | 0.404 | 0.611 |
| gpt3 | 0.511 | **0.557** | 0.472 |
| gpt4 | **0.644** | 0.543 | 0.792 |

**Table 4.** Results for the Fly to Zebrafish anatomy ontology mapping task. As seen for the combined scores (Table 3), Lexmatch achieves high recall but low precision. LogMap offers a demonstrable improvement in precision at the cost of recall, as does our method when used with GPT-3. MapperGPT with GPT-4 attains the highest accuracy by way of recall approaching that of Lexmatch and similar precision to that of the GPT-3-driven approach.

In this task, GPT-4 scored highest in both accuracy and precision.

In the second anatomy ontology matching task, methods were evaluated on their ability to reproduce mappings between the Fly anatomy ontology (FBbt) and the *C. elegans* (roundworm) anatomy ontology (WBbt). A full comparison of the results is provided in **Table 5**, with score distributions in **Figure 2**. Again, MapperGPT with GPT-4 demonstrates the highest F1 score. In this case, however, the most accurate method is that with the highest precision.

| method | F1 | P | R |
|---|---|---|---|
| lexmatch | 0.257 | 0.152 | **0.854** |
| logmap | 0.520 | 0.441 | 0.634 |
| gpt3 | 0.427 | 0.471 | 0.390 |
| gpt4 | **0.660** | **0.585** | 0.756 |

**Table 5.** Results for the Fly to Worm anatomy ontology mapping task. Results are similar to those of the fly vs. zebrafish anatomy mapping, with Lexmatch achieving high recall but very low accuracy. MapperGPT with GPT-4 outperforms the other methods on precision and reaches a high recall score, though its recall does not exceed that of Lexmatch.

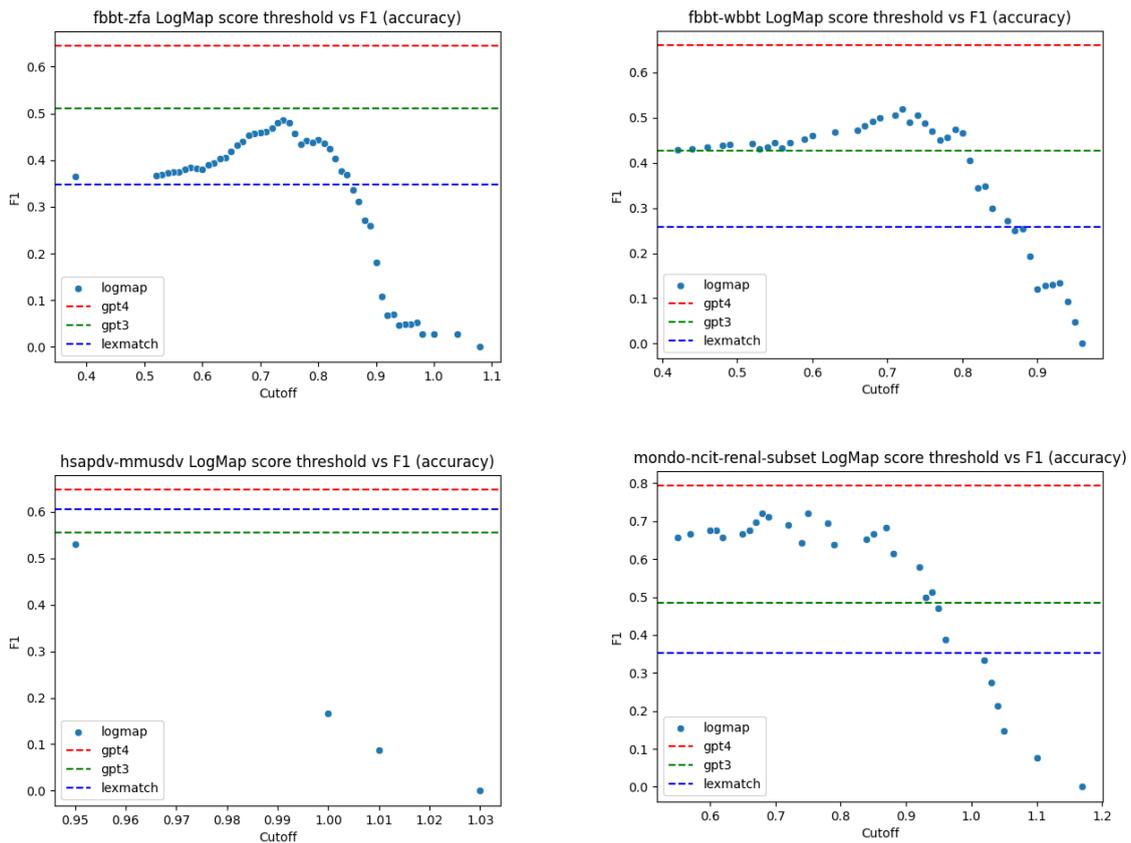

**Figure 2:** Combined LogMap score threshold vs. accuracy of other approaches across individual mapping tasks. Each plot displays results for all four methods in a single mapping task: A) fly vs. zebrafish anatomy term mapping; B) fly vs. worm anatomy term mapping; C) Human vs. mouse developmental stage term mapping; and D) renal disease term mapping. MapperGPT with GPT-4 consistently outperforms all other methods. In both anatomy term mapping tasks, the highest LogMap scores are similar to or exceed those for MapperGPT with GPT-3. For developmental stage mapping, all methods outperform LogMap's highest F1 scores. In renal disease term mapping, the highest F1 scores for LogMap exceed those for MapperGPT with GPT-3 but not the same method with GPT-4.

## Developmental Stage ontology task results

In the developmental stage ontology matching task, methods were evaluated on their ability to reproduce mappings between human developmental stages (from HsapDv) and the corresponding stages for mouse (from MmusDv). Results of this task (**Table 6** and **Figure 2**) resemble those of the anatomy tasks, though with a much larger gap in recall between Lexmatch and our approach, even with the GPT-4 model, and very high precision for MapperGPT with GPT-4.

| method | F1 | P | R |
|---|---|---|---|
| lexmatch | 0.606 | 0.455 | **0.909** |
| logmap | 0.531 | 0.405 | 0.773 |
| gpt3 | 0.556 | 0.714 | 0.455 |
| gpt4 | **0.647** | **0.917** | 0.500 |

**Table 6.** Results for the Human to Mouse developmental stage ontology mapping task. MapperGPT with GPT-4 achieves the highest F1 score, primarily due to high precision. The recall of this approach, however, is exceeded by both Lexmatch and LogMap.

## Disease matching task results

In the disease ontology matching task, methods were evaluated on their ability to reproduce mappings between disease terms concerning kidney disease. Results of this task (**Table 7** and **Figure 2**) show very high recall for most methods, and though our method using GPT-4 achieves the highest F1 score, it is less than 10% higher than LogMap.

| method | F1 | P | R |
|---|---|---|---|
| lexmatch | 0.352 | 0.214 | **1.000** |
| logmap | 0.721 | 0.611 | 0.880 |
| gpt3 | 0.486 | 0.378 | 0.680 |
| gpt4 | **0.793** | **0.697** | 0.920 |

**Table 7.** Results for the renal disease term mapping task. The baseline Lexmatch approach achieves perfect recall in this task, though MapperGPT with GPT-4 comes close while also achieving the highest F1 score across all methods due to its high precision.

## Discussion

In all of the presented challenges, GPT-4 beats all other systems, including the SOTA ontology matching system LogMap, on accuracy (F1 score). While the problem sets are relatively small, they are also particularly difficult for various reasons. For example, cross-species matching problems are hard because of the divergent terminologies used for the matching; this is evident by the relatively poor performance our naive lexical matching baseline (Lexmatch) exhibits throughout the tested problems. Similarly, disease names are known to vary widely across resources.

While GPT-4 emerges as the winner, the performance improvements over SOTA for the disease matching task was modest, which suggests that not a lot of additional background knowledge was taken into account to solve this challenge. Larger test sets and a qualitative analysis are needed to determine for which kinds of problems LLMs have the biggest positive effect.

One potential advantage LLMs already exhibit compared to SOTA matching tools is that they are not as easily confused by domain-specific lexical variation. For example, an LLM is more easily able to understand that "Wilms tumor 1", "Wilms tumor type 1", "Wilms tumor type I" and similar are all lexical variations of the same string, which requires some background knowledge about the word "type" in conjunction with disease names. Secondly, they can utilize background knowledge, not present in the ontologies, providing a notable advantage over traditional methods, such as most SOTA matching tools including LogMap. This characteristic allows the LLM to resemble more of a student intern than a traditional matching tool. Despite these capabilities, our data indicates that even the top-performing method, GPT-4, achieves only about 67% overall accuracy, falling short of the performance attained through mappings by a professional curator. Nonetheless, as language models like LLMs enhance their ability to use context in prompts, they are poised to rapidly narrow this performance gap.

## Limitations of the method

*Proprietary models.* Our best results were achieved using GPT-4. However, at this time, GPT-4 is expensive to run, so we do not recommend its use in cases where simpler lexical methods should suffice. We are exploring open models that can be executed locally.

*Variation across runs*. LLMs are non-deterministic by nature, which means that the results of any given run could differ from any other. While we did run the experiment multiple times to confirm that the overall results (ranking of the best performing approaches) are not affected, we did not perform a formal analysis of variance.

## Future Work

The current implementation of MapperGPT is designed to refine an existing set of candidate mappings. Recall of common, naive lexical approaches for the problems presented here was around 88%, which means that in a real world scenario, we need to expect that at least 10-20% of true positive mappings are missed by such an approach. We are experimenting with the possibility of using LLMs not only in the refinement process (such as presented here) but also in the process of proposing suitable mapping candidates. This is a very different problem and will require the LLMs to directly access the ontology using Retrieval Augmented Generation (RAG). We are also exploring the use of RAG to provide the most relevant gold-standard mappings as in-context examples.

We are integrating MapperGPT into our Boomer[20] pipeline to make BoomerGPT (https://github.com/monarch-initiative/boomer-gpt), a hybrid neuro-symbolic mapping tool that integrates probabilistic inference, description logic reasoning, lexical methods, rule-based methods, and LLMs for the problem of merging diverse ontologies into a single, coherent whole.

# Conclusions

We are in the early stages of exploring the use of LLMs for semantic mapping problems, but given the promising performance that current LLMs already exhibit on complex matching problems, it is likely that this field is going to progress fast. In this study we confirm that the current generation of LLMs can perform better than SOTA ontology matching tools on a diverse set of ontology matching problems. However, in their current form, the approach is with a 67% overall accuracy still considerably behind human curator performance. Our reference implementation, MapperGPT, is integrated into the OntoGPT framework that leverages LLMs for the construction of ontologies.

# Acknowledgments

This work was made possible by the contributions of countless free software developers, upon which much of our modern infrastructure is built.

# Funding

This work was supported in part by the Office of the Director National Institute of Health 5R24OD011883-12 and 3R24OD011883-11S1, and NHGRI Center of Excellence in Genome Sciences RM1 HG010860. JR, SC, SM, HH, NLH, CJM, and JHC were supported in part by the Director, Office of Science, Office of Basic Energy Sciences, of the U.S. Department of Energy under Contract No. DE-AC02-05CH11231. We also gratefully acknowledge Bosch Research for their support of this research project.

1. Ghazvinian A, Noy NF, Musen MA. Creating mappings for ontologies in biomedicine: simple methods work. AMIA Annu Symp Proc. 2009 Nov 14;2009:198–202. https://www.ncbi.nlm.nih.gov/pubmed/20351849 PMCID: PMC2815474

2. Euzenat J, Meilicke C, Stuckenschmidt H, Shvaiko P, Trojahn C. Ontology Alignment Evaluation Initiative: Six Years of Experience. In: Spaccapietra S, editor. Journal on Data Semantics XV. Berlin, Heidelberg: Springer Berlin Heidelberg; 2011. p. 158–192. https://doi.org/10.1007/978-3-642-22630-4_6

3. Jiménez-Ruiz E, Cuenca Grau B. LogMap: Logic-Based and Scalable Ontology Matching. The Semantic Web – ISWC 2011. Springer Berlin Heidelberg; 2011. p. 273–288. http://dx.doi.org/10.1007/978-3-642-25073-6_18

4. Wang LL, Bhagavatula C, Neumann M, Lo K, Wilhelm C, Ammar W. Ontology Alignment in the Biomedical Domain Using Entity Definitions and Context. arXiv [cs.CL]. 2018. http://arxiv.org/abs/1806.07976

5. Kolyvakis P, Kalousis A, Kiritsis D. DeepAlignment: Unsupervised ontology matching with refined word vectors. Proceedings of the 2018 Conference of the North American Chapter of the Association for Computational Linguistics: Human Language Technologies, Volume 1 (Long Papers). Stroudsburg, PA, USA: Association for Computational Linguistics; 2018. https://arodes.hes-so.ch/record/2623/files/Kalousis_2018_DeepAlignment.pdf

6.  Iyer V, Agarwal A, Kumar H. Multifaceted Context Representation using Dual Attention for Ontology Alignment. arXiv [cs.AI]. 2020. http://arxiv.org/abs/2010.11721

7.  Amir M, Baruah M, Eslamialishah M, Ehsani S, Bahramali A, Naddaf-Sh S, Zarandioon S. Truveta Mapper: A Zero-shot Ontology Alignment Framework. arXiv [cs.LG]. 2023. http://arxiv.org/abs/2301.09767

8.  He Y, Chen J, Dong H, Horrocks I. Exploring Large Language Models for Ontology Alignment. arXiv [cs.AI]. 2023. http://arxiv.org/abs/2309.07172

9.  Matentzoglu N, Balhoff JP, Bello SM, Bizon C, Brush M, Callahan TJ, Chute CG, Duncan WD, Evelo CT, Gabriel D, Graybeal J, Gray A, Gyori BM, Haendel M, Harmse H, Harris NL, Harrow I, Hegde HB, Hoyt AL, Hoyt CT, Jiao D, Jiménez-Ruiz E, Jupp S, Kim H, Koehler S, Liener T, Long Q, Malone J, McLaughlin JA, McMurry JA, Moxon S, Munoz-Torres MC, Osumi-Sutherland D, Overton JA, Peters B, Putman T, Queralt-Rosinach N, Shefchek K, Solbrig H, Thessen A, Tudorache T, Vasilevsky N, Wagner AH, Mungall CJ. A Simple Standard for Sharing Ontological Mappings (SSSOM). Database . 2022 May 25;2022. http://dx.doi.org/10.1093/database/baac035 PMCID: PMC9216545

10. Costa M, Reeve S, Grumbling G, Osumi-Sutherland D. The Drosophila anatomy ontology. J Biomed Semantics. 2013 Oct 18;4(1):32. http://dx.doi.org/10.1186/2041-1480-4-32 PMCID: PMC4015547

11. Van Slyke CE, Bradford YM, Westerfield M, Haendel MA. The zebrafish anatomy and stage ontologies: representing the anatomy and development of Danio rerio. J Biomed Semantics. 2014 Feb 25;5(1):12. http://dx.doi.org/10.1186/2041-1480-5-12 PMCID: PMC3944782

12. Mungall C, Harshad, Kalita P, Hoyt CT, Patil S, Joachimiak M p., Flack J, Linke D, Harris N, Caufield H, Reese J, Moxon S, Rutherford K, Matentzoglu N, Deepak, Glass, de Souza V, Jacobsen J, Schaper K, Lera-Ramirez M, Tan S, Lubiana T. INCATools/ontology-access-kit: v0.5.20. 2023. https://zenodo.org/record/8371372

13. Jackson R, Matentzoglu N, Overton JA, Vita R, Balhoff JP, Buttigieg PL, Carbon S, Courtot M, Diehl AD, Dooley DM, Duncan WD, Harris NL, Haendel MA, Lewis SE, Natale DA, Osumi-Sutherland D, Ruttenberg A, Schriml LM, Smith B, Stoeckert CJ Jr, Vasilevsky NA, Walls RL, Zheng J, Mungall CJ, Peters B. OBO Foundry in 2021: operationalizing open data principles to evaluate ontologies. Database . 2021 Oct 26;2021. http://dx.doi.org/10.1093/database/baab069 PMCID: PMC8546234

14. Noy NF, Shah NH, Whetzel PL, Dai B, Dorf M, Griffith N, Jonquet C, Rubin DL, Storey M-A, Chute CG, Musen MA. BioPortal: ontologies and integrated data resources at the click of a mouse. Nucleic Acids Res. 2009 Jul;37(Web Server issue):W170–3. http://dx.doi.org/10.1093/nar/gkp440 PMCID: PMC2703982

15. Jackson RC, Balhoff JP, Douglass E, Harris NL, Mungall CJ, Overton JA. ROBOT: A Tool for Automating Ontology Workflows. BMC Bioinformatics. 2019 Jul 29;20(1):407. http://dx.doi.org/10.1186/s12859-019-3002-3 PMCID: PMC6664714

16. Caufield JH, Hegde H, Emonet V, Harris NL, Joachimiak MP, Matentzoglu N, Kim H, Moxon SAT, Reese JT, Haendel MA, Robinson PN, Mungall CJ. Structured prompt interrogation and recursive extraction of semantics (SPIRES): A method for populating knowledge bases using zero-shot learning. arXiv [cs.AI]. 2023. http://arxiv.org/abs/2304.02711

17. Vasilevsky NA, Matentzoglu NA, Toro S, Flack JE IV, Hegde H, Unni DR, Alyea GF, Amberger JS, Babb L, Balhoff JP, Bingaman TI, Burns GA, Buske OJ, Callahan TJ, Carmody LC, Cordo PC, Chan LE, Chang GS, Christiaens SL, Dumontier M, Failla LE, Flowers MJ, Garrett HA Jr, Goldstein JL, Gration D, Groza T, Hanauer M, Harris NL, Hilton JA, Himmelstein DS, Hoyt CT, Kane MS, Köhler S, Lagorce D, Lai A, Larralde M, Lock A, López Santiago I, Maglott DR, Malheiro AJ, Meldal BHM, Munoz-Torres MC, Nelson TH, Nicholas FW, Ochoa D, Olson DP, Oprea TI, Osumi-Sutherland D, Parkinson H, Pendlington ZM, Rath A, Rehm HL, Remennik L, Riggs ER, Roncaglia P, Ross JE, Shadbolt MF, Shefchek KA, Similuk MN, Sioutos N, Smedley D, Sparks R, Stefancsik R, Stephan R, Storm AL, Stupp D, Stupp GS, Sundaramurthi JC, Tammen I, Tay D, Thaxton CL, Valasek E, Valls-Margarit J, Wagner AH, Welter D, Whetzel PL, Whiteman LL, Wood V, Xu CH, Zankl A, Zhang XA, Chute CG, Robinson PN, Mungall CJ, Hamosh A, Haendel MA. Mondo: Unifying diseases for the world, by the world. bioRxiv. 2022. https://www.medrxiv.org/content/10.1101/2022.04.13.22273750.abstract

18. Diehl AD, Meehan TF, Bradford YM, Brush MH, Dahdul WM, Dougall DS, He Y, Osumi-Sutherland D, Ruttenberg A, Sarntivijai S, Van Slyke CE, Vasilevsky NA, Haendel MA, Blake JA, Mungall CJ. The Cell Ontology 2016: enhanced content, modularization, and ontology interoperability. J Biomed Semantics. 2016 Jul 4;7(1):44. http://dx.doi.org/10.1186/s13326-016-0088-7 PMCID: PMC4932724

19. Mungall CJ, Torniai C, Gkoutos GV, Lewis SE, Haendel MA. Uberon, an integrative multi-species anatomy ontology. Genome Biol. 2012 Jan 31;13(1):R5. http://dx.doi.org/10.1186/gb-2012-13-1-r5 PMCID: PMC3334586

20. boomer: Bayesian OWL ontology merging. Github; https://github.com/INCATools/boomer